\documentclass[runningheads]{llncs}
 
\usepackage{eccv}

\usepackage{eccvabbrv}

\usepackage{graphicx}
\usepackage{booktabs}

\usepackage[accsupp]{axessibility}  

\usepackage{hyperref}

\usepackage{orcidlink}
\usepackage{url}
\usepackage{multirow}
\usepackage{changes}

\begin{document}

\title{Unsupervised Multi-modal Medical Image Registration via Invertible Translation} 

\titlerunning{Medical Image Registration via Invertible Translation}

\author{Mengjie Guo\inst{1,2}, Jinming Duan\inst{1,3*}}

\authorrunning{M. Guo}

\institute{University of Birmingham, Birmingham, UK 
\and Southern University of Science and Technology, Shenzhen, China
\and University of Manchester, Manchester, UK\\
\email{jinming.duan@manchester.ac.uk}\\
}

\maketitle

\begin{abstract}
In medical imaging, the alignment of multi-modal images plays a critical role in providing comprehensive information for image-guided therapies. Despite its importance, multi-modal image registration poses significant challenges due to the complex and often unknown spatial relationships between different image modalities. To address this, we introduce a novel unsupervised translation-based multi-modal registration method, termed Invertible Neural Network-based Registration (INNReg). INNReg consists of an image-to-image translation network that converts multi-modal images into mono-modal counterparts and a registration network that uses the translated mono-modal images to align the multi-modal images. Specifically, to ensure the preservation of geometric consistency after image translation, we introduce an Invertible Neural Network (INN) that leverages a dynamic depthwise convolution-based local attention mechanism. Additionally, we design a novel barrier loss function based on Normalized Mutual Information to impose constraints on the registration network, which enhances the registration accuracy. The superior performance of INNReg is demonstrated through experiments on two public multi-modal medical image datasets, including MRI T1/T2 and MRI/CT pairs. The code is available at \url{https://github.com/MeggieGuo/INNReg}.


\end{abstract}

\section{Introduction}
\label{sec:intro}

Multi-modal image registration aims at aligning images from different modalities into a common coordinate system, facilitating accurate comparison, integration, and analysis. This process is particularly critical in the fusion of medical images from distinct modalities such as Magnetic Resonance Imaging (MRI), Positron Emission Tomography (PET), Computed Tomography (CT), etc. 
In medical imaging, each modality presents unique insights into anatomical structures by highlighting various physical attributes, and the fusion across modalities not only leverages the unique strengths of each modality but also provides a more holistic understanding of patient anatomy, significantly enhancing diagnostic accuracy and the effectiveness of treatment strategies.

The challenge of multi-modal image registration lies in bridging the gap between distinct image characteristics inherent to each modality. Early works for addressing this challenge mainly use supervised learning methods that rely on labelled data, such as ground-truth deformation fields or segmentation masks \cite{simonovsky2016deep}. However, the applicability of these supervised learning methods is severely limited by the scarcity of annotated data due to the heavy effort required for manual labelling. Particularly, for registration, it is difficult to acquire human-labelled deformation fields. Due to these reasons, many research \cite{de2017end, mahapatra2018joint, cao2018deep, mahapatra2018deformable} instead focus on unsupervised learning strategies and attempt to design proper qualitative similarity metrics on the alignment between images from different modalities. These methods eliminate the need for labelled data, though their performance heavily relies on the design of similarity metrics, which can be challenging to define. 

For instance, metrics like the Sum of Squared Differences (SSD) and Normalized Cross-Correlation (NCC) are unsuitable for multi-modal contexts, often resulting in suboptimal registration outcomes \cite{balakrishnan2019voxelmorph, cao2018deep}. The Modality-Independent Neighbourhood Descriptor (MIND) focuses on local structural information and exhibits poor performance in global alignment tasks \cite{heinrich2012mind}. On the other hand, while Normalized Mutual Information (NMI) fares better in global alignment, it struggles with local alignment tasks \cite{mahapatra2018joint}.

Recent works on multi-modal image registration have introduced a new deep learning-based paradigm. This paradigm typically consists of a Generative Adversarial Network (GAN) \cite{isola2017image} that translates images of different modalities into the same modality, alongside a mono-modal image registration network for aligning the translated images. It is noteworthy that within this paradigm, the objective is for the GAN to harmonize the style of different modalities after translation while preserving the underlying geometry. Subsequently, the registration network learns the deformation field and warps the image geometry accordingly. In contrast to the aforementioned unsupervised learning strategies reliant on similarity metrics, translation-based registration methods offer a workaround to the challenge of devising effective metrics. However, they frequently introduce geometric inconsistencies, leading to a notable decline in registration performance and potential mode collapse when employing GANs in the translation process \cite{arar2020unsupervised}.

In this paper, we propose a novel unsupervised translation-based multi-modal registration method, termed Invertible Neural Network-based Registration (INNReg). Unlike most existing translation-based methods that rely on GANs, INNReg utilizes an Invertible Neural Network (INN) as the translation network. The invertible nature of INN facilitates the learning of modality mapping while ensuring geometric consistency. Additionally, to enhance feature extraction during the translation process, we augment INN with a dynamic depthwise convolution (DDWConv) based on a local attention scheme, which encourages the network to focus on specific regions of the source image \cite{han2021connection}. Furthermore, we propose a novel \emph{barrier} NMI loss function to impose constraints on the registration network, addressing the issue of low registration accuracy caused by geometric inconsistency in the translation. To assess the effectiveness of our proposed method, we conduct extensive experiments on two public multi-modal medical image datasets, including MRI T1/T2 and MRI/CT image pairs. Both qualitative and quantitative analyses demonstrate the competitive performance of our INNReg method in multi-modal registration 

The main contributions of our work can be summarized as follows: 
\begin{itemize}
\item We propose INNReg, the first multi-modal image registration method that integrates a translation task through INN. Notably, it effectively addresses the inherent issue of geometric inconsistency found in previous multi-modal image registration approaches that incorporate image translation. In these methods, the image after translation alters not only the appearance but also the underlying geometry of the original image.

\item We integrate local attention through dynamic depthwise convolution to capture fine-grained details during the translation process. This method further ensures the preservation of geometric consistency during image translation,
\item We propose a novel barrier NMI loss function to constrain the registration network, effectively circumventing the inherent limitations of NMI and thereby improving registration accuracy.

\item Through extensive experiments, we demonstrate the superior qualitative and quantitative performance of INNReg compared to state-of-the-art multi-modal image registration approaches. 

\end{itemize}

\section{Related Work}

\subsection{I2I Translation-based Multi-modal Image Registration}

In recent years, multi-modal image registration has seen the emergence of several unsupervised translation-based methods. These methods generally adhere to a registration-by-translation paradigm, wherein an Image-to-Image (I2I) translation network is tasked with synthesizing images that mimic the target modality's appearance. This allows for the application of mono-modality similarity metrics to multi-model registration tasks. Among unsupervised translation techniques, cycle-consistent generative adversarial networks (CycleGAN \cite{zhu2017unpaired}) have gained popularity due to their ability to enforce content preservation through cycle consistency. However, this approach often generates multiple solutions, risking the anatomical accuracy of translated images by introducing artefacts and potentially degrading the quality of multi-modal registration \cite{kong2021breaking}.

To avoid these challenges, various strategies have been explored beyond the scope of CycleGAN. Qin et al. \cite{qin2019unsupervised} leveraged image disentanglement to segregate images into domain-invariant shape features and domain-specific appearance features, facilitating the training of a registration network using the shape features across modalities. However, it is challenging to clearly define and constrain domain-invariant and domain-specific features between different modalities. Arar et al. \cite{arar2020unsupervised} proposed a methodology aiming to make the translation and registration steps commutative, thereby indirectly promoting the structural consistency of the translation network. However, the reliance on a GAN-based framework implies that structural consistency could be compromised by the discriminator's influence.

\subsection{Invertible Neural Network}

Invertible Neural Network (INN) \cite{dinh2014nice, dinh2016density} has several advantages over GAN, which are commonly used in conventional forward propagation methods. Firstly, GAN often faces significant information loss, while INN only results in slight information loss benefit from both forward and backward computations. Secondly, INN has a significantly lower risk of mode collapse, which is a common issue in GAN training. These advantages of INN make it particularly suited for applications such as image fusion \cite{zhou2022effective, zhao2023cddfuse}, image denoising \cite{huang2022winnet} and I2I translation \cite{ardizzone2021conditional}. For example, Zhao et al. \cite{zhao2023cddfuse} utilize the INN blocks for lossless information transmission in the encoder of their image fusion model. Ardizzone et al. \cite{ardizzone2021conditional} combine an INN with an unconstrained feed-forward network for conditioning to address the task of diverse I2I translation for natural images. Huang et al. \cite{huang2022winnet} propose a novel wavelet-inspired invertible network with redundant invertible sparsifying transforms for image denoising.
The reversible architecture of INNs enables precise manipulation of latent spaces, facilitating complex generative tasks while guaranteeing accurate reconstruction of the original input data. INNs have also shown significant promise in enhancing image quality by establishing a direct correlation between inputs and their high-resolution outputs. Despite these advancements, the potential of INNs in the realm of I2I translation remains untapped.

\subsection{Local Attention by Dynamic Depthwise Convolution}
Attention mechanisms \cite{vaswani2017attention} are commonly-used techniques in the fields of image processing \cite{parmar2018image, zhang2019self} and computer vision \cite{hu2018squeeze, wang2018non} for encoding long-range dependency in extracted features. In I2I translation, the integration of attention mechanisms has emerged as a significant advancement, enabling more focused and context-aware transformations between source and target domains. To list a few, Tang et al. \cite{tang2019attention} proposed an Attention-Guided Generative Adversarial Network (AGGAN) to transfer high-level semantic parts of images to obtain high-quality images; Tang et al. \cite{tang2022multi} developed a Multi-Channel Attention Selection GAN to follow external semantic guidance for I2I translation. 

In vision tasks, attention usually acts as a dynamic information aggregator in spatial and temporal domains. Specifically, the local attention mechanism forms the keys and values in a window that the query lies in. The attention output is the weighted aggregation of the corresponding values in the local window, where the weights are the softmax normalisation of the dot-product between the queries and the keys. A notable work in this category is \cite{han2021connection}, which extensively analyses local attention, and establishes the connection between dynamic depthwise convolution and local self-attention by connecting the attention weights for self-attention and the dynamic weights for convolution. Moreover, they empirically observe that the local attention models based on dynamic depthwise convolution perform on par with or slightly better than previous implementations but have lower computational complexity.

\section{Methodology}

\begin{figure}[tb]
  \centering
  \includegraphics[width=\textwidth]{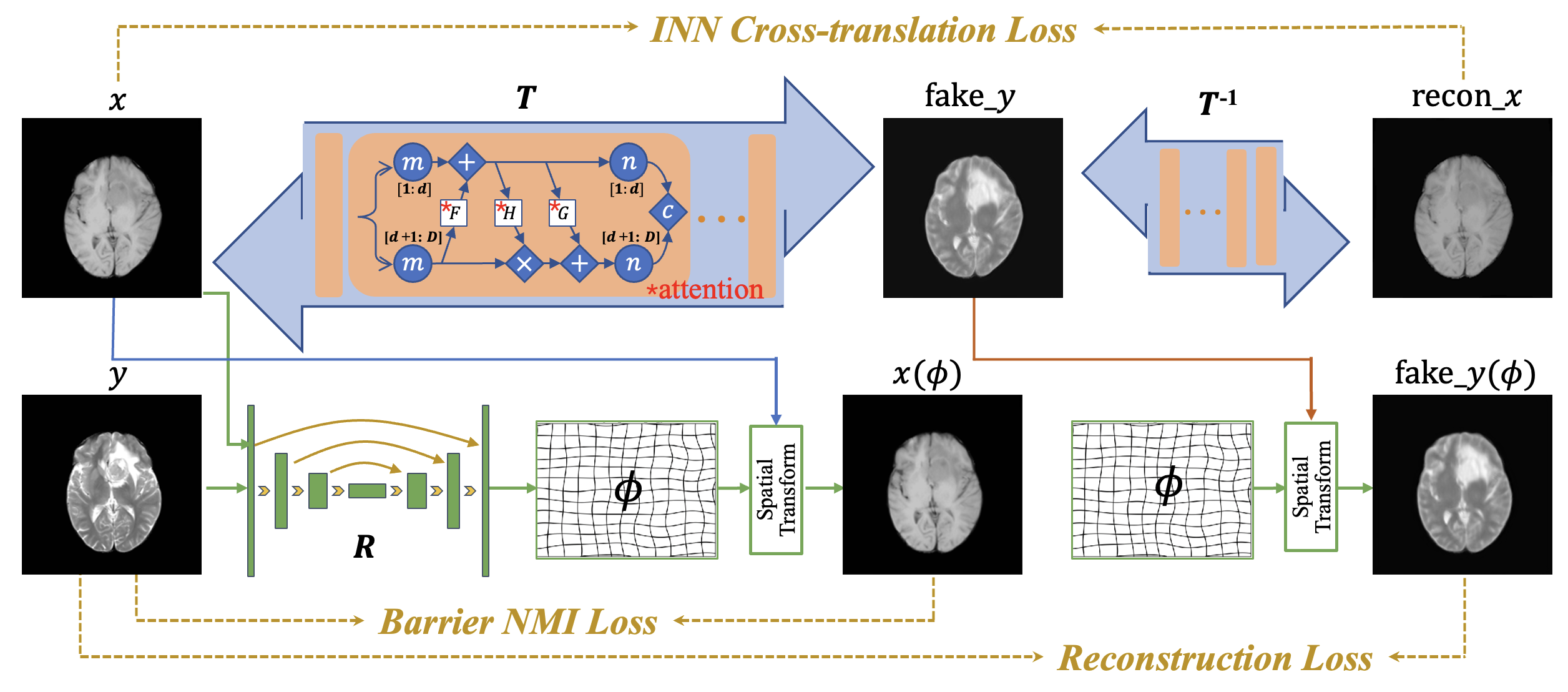}
  \caption{ The framework of the proposed INNReg for multi-modal image registration. In the translation process (\textbf{Top}), the forward transformation $T$ translates the source image $x$ into the domain of $y$ as $\text{fake}\_y$, and the reverse transformation $T^{-1}$ translates $\text{fake}\_y$ back into the domain of $x$ as $\text{recon}\_x$. The registration process (\textbf{Bottom}) first uses the registration network $R$ to predict the deformation field $\phi$ from $x$ and $y$, and then uses $\phi$ and a Spatial Transform function to warp $x$ and $\text{fake}\_y$, resulting in $x(\phi)$ and $\text{fake}\_y(\phi)$, respectively.
  }
  \label{fig_outline}
\end{figure}

We propose a translation-based multi-modal registration method, referred to as Invertible Neural Network-based Registration (INNReg). INNReg includes an invertible I2I translation network for translating images across various modalities and a deformable registration network for aligning multi-modal medical images based on the translated images. 
We introduce the registration network in Section \ref{sec_registration}, and describe the invertible translation network in Section \ref{sec_INN}. To enhance feature extraction of the translation network, we equip it with a dynamic depthwise convolution-based local attention scheme. The losses in the proposed INNReg are detailed in Section \ref{sec_loss}. The pipeline of our method is depicted in Figure \ref{fig_outline}. Note that the registration and translation networks are trained jointly, while only the registration network is used in the test.

\subsection{Registration Model} \label{sec_registration}

The registration model first employs a network $R$ to learn the deformation field $\phi = R(x,y)$ from a pair of images $(x, y)$ with distinct modalities. The field $\phi$ is a matrix of 2D vectors, where each vector indicates the displacement for aligning individual pixels from the source image $x$ with the structure of the target image $y$. With $\phi$ in hand, we warp $x$ to align with $y$ by a spatial transform function. The learned $\phi$ is domain-free, meaning that it applies to both the domain $\mathcal{X}$ of the source images and $\mathcal{Y}$ of the target images.

\subsection{Translation Model} \label{sec_INN}

We aim to design a translation model in INNReg to harmonize image styles from different modalities while preserving the crucial geometry details inherent to each image, which is particularly vital for translation-based multi-modal image registration methods. To achieve this, we develop the translation model based on INN, which allows for mutual generations of features and therefore ensures a high fidelity of information preservation between its input and output.

The INN in our translation model can be described as a function \(T: \mathcal{X} \rightarrow \mathcal{Y}\) mapping from the source image domain $\mathcal{X}$ to the target image domain $\mathcal{Y}$ and consists of a stack of Invertible Blocks (InvBlocks) $\{T_i\}_{i=1}^k$. Specifically, for any source image \(x\), its translation via $T$ is
\begin{align}
    \text{fake}\_y=T(x) &= T_{0} \circ T_{1} \circ T_{2} \circ \cdots \circ T_{k}(x),
\end{align}
and the inverse translation back to the source image domain $\mathcal{X}$ is
\begin{align}
    \text{recon}\_x &= T^{-1}(\text{fake}\_y)= T_{k}^{-1} \circ T_{k-1}^{-1} \circ \cdots \circ T_{0}^{-1}(\text{fake}\_y),
\end{align}
where $T^{-1}:\mathcal{Y}\rightarrow\mathcal{X}$ is the inverse of $T$ and each $T_i^{-1}$ is the inverse of the InvBlock $T_i$.

In our method, each InvBlock $T_i$ is realized through an affine coupling layer \cite{dinh2016density}. Given a $D$-dimensional vector $m$ and an integer $d\in [1,D)$, the layer first splits $m$ into two vectors $m_{1:d} $ and $m_{d+1:D}$, and then performs the additive affine transformations:
\begin{align}
n_{1:d} &= m_{1:d}+F\left ( m_{d+1:D} \right )\label{eq:n1d}\\
n_{d+1:D} &= m_{d+1:D} \odot \exp \left ( H\left ( n_{1:d} \right )  \right ) +G\left ( n_{1:d} \right ) \label{eq:ndD}
\end{align}
where $\odot$ is the Hadamard product, and $F$, $H$, and $G$ are arbitrarily complicated functions of $m$ and not necessarily invertible. In our implementation, the three functions $F$, $H$, and $G$ are realized by a common densely connected convolutional block, referred to as DenseBlock in \cite{song2020efficient, wang2018esrgan}. Note that given $n_{1:d}$ and $n_{d+1:D}$, the inverse transformations of \eqref{eq:n1d}--\eqref{eq:ndD} can be computed as
\begin{align}
m_{d+1:D} &= (n_{d+1:D}-G\left ( n_{1:d} \right ) )\odot \exp \left( -H\left( n_{1:d}\right) \right) \\
m_{1:d} &=  n_{1:d} - F\left(m_{d+1:D}\right).
\end{align}

\subsubsection{Local Attention by DDWConv.}
To improve the feature extraction ability of the translation model, we equip INN with a dynamic depthwise convolution (DDWConv)-based local attention scheme. Local attention mechanisms are particularly useful in I2I translation tasks, in the sense that they can help the model refine the details in the generated output by focusing on specific regions of the input image.

Unlike existing local attention schemes that are mainly designed based on feature aggregation modules with Key, Query, and Value, we adapt DDWConv to realize local attention. The recent work \cite{han2021connection} points out that to yield excellent performance, local attention schemes should satisfy three principles: sparse connectivity, weight sharing, and dynamic weighting. These characteristics also exist in dynamic depthwise convolution in CNN, and compared to common local attention realizations, the computational cost of using DDWConv is often lower. Inspired by this, we realize local attention by DDWConv. To illustrate DDWConv, we first introduce depthwise convolution:
\begin{align}\label{eq:DC}
DepthwiseConv_{(i,j)} &=\sum_{p,q}^{P,Q} W_{\left (p,q\right ) }\cdot x_{\left ( i+p,j+q\right ) },
\end{align}
where $(i,j)$ is a coordinate, $P$ and $Q$ are the height and width of $x$, respectively, $W$ represents weight, and $x$ is an input feature map. In DDWConv, the weight $W$ in \eqref{eq:DC} is learned \emph{dynamically} by aggregating multiple convolution kernels $\pi_kW_k$, $k=1,\ldots,K$:
\begin{align}
W(x)  &= \sum_{k= 1}^{K} \pi_{k}(x)W_k.
\end{align}

\begin{figure}[h]
  \centering
  \includegraphics[width=0.6\textwidth]{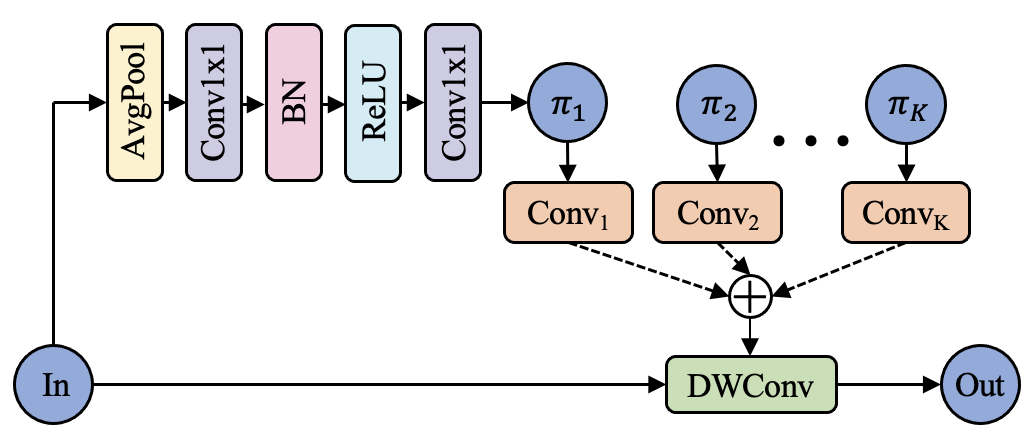}
  \caption{DDWConv-based local attention.
  }
  \label{fig_attention}
\end{figure}

Note that different from the (static) depthwise convolution in which $W$ is fixed, the weight $W(x)$ in DDWConv varies for each input feature map $x$. Due to this feature, DDWConv suits the input's characteristics and, therefore, can potentially extract more relevant and discriminative features from the input data, which further leads to the superior performance of local attention in feature extraction. A sketch of the DDWConv-based local attention scheme is provided in Fig. \ref{fig_attention}.

\subsection{Loss Functions} \label{sec_loss}

The proposed INNReg is trained in an end-to-end manner. The total loss function consists of four components, including a pixel-wise reconstruction loss, a barrier NMI registration loss, an INN cross-translation loss, and an additional smooth loss. During the training, we use the barrier NMI registration loss to constrain the registration network, the INN cross-translation loss to optimize the translation network, the pixel-wise reconstruction loss to jointly optimize the registration network and the translation network to generate outputs that closely match the target image, and the smooth loss to preserve the smoothness of the predicted deformation field.

\subsubsection{Pixel-wise Reconstruction Loss.}
The pixel-wise reconstruction loss is the key to the success of the translation-based methods in handling multi-modal registration problems. In our method, it is the $\ell_1$-reconstruction loss between the generated results (i.e., $\text{fake}\_y(\phi)$) and the target image $y$:
\begin{align}
\mathcal{L}_{\text{recon}}\left ( T,R \right ) & = \left \| \text{fake}\_y\left ( \phi \right ) -y \right \| _1,\label{eq:recon_loss}
\end{align}
where $T$ and $R$ are the translation and the registration network, respectively. The use of the loss \eqref{eq:recon_loss} enforces $\text{fake}\_y(\phi)$ to approximate the target image $y$.

\subsubsection{Barrier NMI Registration Loss.} The barrier NMI registration loss is proposed to improve the accuracy of the registration network. In the translation-based multi-modal registration methods, the deformation in the translation network often deteriorates the accuracy of the registration network. Our idea of addressing this issue is to constrain the similarity between the warped source image and the target image.

In our method, we adopt the widely used Normalized Mutual Information (NMI) to measure the similarity between two images by quantifying their mutual information of intensity values. The NMI metric between the warped source image $x(\phi)$ and the target image $y$ can be expressed as
\begin{align}
I_{\text{NMI}} \left(y , x(\phi)  \right)=\frac{H\left ( y \right ) +H\left ( x (\phi)   \right ) }{M\left (y , x (\phi)   \right ) },
\end{align}
where $H\left ( y\right ) $ and $H\left ( x (\phi)  \right ) $ denote the marginal entropies of the source and the target image, respectively, and $M\left (y, x(\phi)   \right) $ is the mutual information between $x(\phi)$ and $y$. Although NMI can also be used as a loss function in DL models for multi-modal image registration, it comes with certain disadvantages \cite{pluim2003mutual}. 
In particular, they cannot effectively capture complex or nonlinear relationships in multi-modal data, potentially limiting the performance of the multi-modal registration network. 

Based on NMI, we propose the following barrier loss function:
\begin{align}\label{eq:barrier}
\mathcal{L}_{\text{barrier}}(I_{\text{NMI}},b)=-\log(b-I_{\text{NMI}}),
\end{align}
where $b$ is a manually set threshold. The barrier function value tends to infinity when $I_{\text{NMI}}$ approaches $b$ and, therefore, adding the barrier loss to the total loss \emph{rigorously} guarantees $I_{\text{NMI}}<b$. This property helps to avoid the accuracy degradation of the registration network caused by deformation in the translation network. To see this, note that if a severe deformation occurs in the translation network, then the registration network will generate an inaccurate deformation field, which further leads to a huge barrier loss. Additionally, since the barrier function value changes slowly when $I_{\text{NMI}}$ is not too close to the threshold $b$, the influence of this barrier loss diminishes, ensuring that the limitations of NMI do not significantly impact the registration outcome.

\subsubsection{INN Cross-Translation Loss.}

One major challenge in I2I translation tasks is the requirement of paired images. For unpaired data, previous methods typically employ the GAN loss for translation tasks \cite{chen2022unsupervised, liu2023geometry}, but it often results in mode collapse. In this method, we introduce an INN cross-translation loss as a constraint of our translation model, which takes the following form:
\begin{align}\label{eq:LINN}
\mathcal{L}_{\text{INN}}(T) & = \left \| \text{recon}\_x -x \right \| _1
\end{align}
where $x$ is the source image, $T$ is the INN translation network, and $\text{recon}\_x=T^{-1}\circ T\left ( x \right )$ with $T^{-1}$ being the reverse transformation of $T$. The reversibility of INN ensures that the translation process using \eqref{eq:LINN} is coherent and maintains the critical attributes of the original images, even in the absence of paired training samples. 
An intuitive explanation is that the INN cross-translation loss preserves the content of the image and only changes the domain-specific attributes.

\subsubsection{Smooth Loss.}  
In deformable registration, the deformation field often suffers from the issue of over-distortion, which makes it hard to establish a precise match between two unaligned images. To prevent over-distortion, we use a smooth loss to restrict the gradients of the predicted deformation field $\phi$:
\begin{align}
\mathcal{L}_{\text {smooth }} & = \frac{1}{N} \sum_{p}\|\nabla \phi(p)\|_2^2,
\end{align}
where each $p$ is a pixel in $\phi$ and $N$ represents the total number of pixels. 

\subsubsection{Total Loss.}
With the above four loss functions, the total loss function for training takes the following form:
\begin{align}
\mathcal{L}_{\text{Total}} & = \lambda_\alpha  \mathcal{L}_{\text{recon}} + \lambda_\beta  \mathcal{L}_{\text{barrier}}+ \lambda_\gamma  \mathcal{L}_{\text{INN}}+ \lambda_\delta  \mathcal{L}_{\text{smooth}}
\end{align}
where $ \lambda_\alpha,  \lambda_\beta, \lambda_\gamma \text{ and } \lambda_\delta$ are the weights to each loss term. 
The purpose of this loss is to improve the multi-modal registration performance by letting the translation network focus on the modality mapping and the registration network on geometry warping.

\section{Experiments}

\subsection{Experimental Settings}

\subsubsection{Datasets.}
We evaluate the registration performance of our INNReg method on two publicly accessible multi-modal medical image datasets, including a T1/T2 weighted MRI image dataset and an MRI/CT image dataset. 

The T1/T2 dataset is from the Brain Tumour Segmentation (BraTS) 2023 Challenge \cite{kazerooni2023brain} containing segmentation labels and is widely used for medical image registration \cite{liu2023geometry, xu2023murf}. The labels delineated three clinical tumour regions of interest (ROIs) within the brain and were approved by expert neuroradiologists. The dataset consists of 60 scan pairs with labels, where each scan is a 3D volume of 240 × 240 × 155. In our experiments, we randomly divide the 60 pairs into 54/6 for training/testing, and in each volume, we extract the middle 50 slices as our input, where each slice is a 2D image of the size $240\times 240$.

The MRI/CT dataset is from Harward \cite{Harward}, and in our experiment, we use 920 image pairs for training and 92 for testing. Since the multi-modal image pairs in the aforementioned two datasets are all aligned, we need to synthetically generate unaligned images from them to train the network. Specifically, we adopt the B-spline transformation method used in \cite{deng2023interpretable, klein2008optimisation} to generate unaligned T1/T2 weighted MRI and MRI/CT pairs. All images used in the experiments are resized as $192\times192$.

\setlength{\tabcolsep}{3.5pt}
\begin{table}[t]
  \caption{The multi-modal image registration results on T1/T2 and MRI/CT dataset. The best and the second best results are marked in \textbf{bold} and by asterisk*, respectively.}
  \label{tab:compare}
  \centering
\begin{tabular}{cc|ccccc}
\hline
\multicolumn{2}{c|}{Methods}       & VM     & Arar et al.     & RegGAN      & Chen et al.   & INNReg(Ours)            \\ \hline
\multicolumn{1}{c|}{\multirow{5}{*}{T1/T2}} & SSIM(\%)↑   & 89.414  & 83.813         & 89.971*           & 89.386  & \textbf{90.555} \\
\multicolumn{1}{c|}{}                           & NCC(\%)↑    & 98.03   & 95.309         & 98.238           & 98.889*  & \textbf{98.746} \\
\multicolumn{1}{c|}{}                           & Dice(\%)↑   & 86.984*  & 83.296         & 84.643           & 81.274  & \textbf{87.117} \\
\multicolumn{1}{c|}{}                           & HD95↓       & 128.318 & 101.806        & \textbf{100.660} & 104.473 & 101.371*        \\
\multicolumn{1}{c|}{}                           & smooth(\%)↓ & 0.238   & \textbf{0.007} & 0.232            & 5.303   & 0.101*          \\ \hline
\multicolumn{1}{c|}{\multirow{3}{*}{MRI/CT}}    & SSIM(\%)↑   & 54.74   & 64.791         & 71.637*           & 63.308  & \textbf{73.067} \\
\multicolumn{1}{c|}{}                           & NCC(\%)↑    & 55.903  & 83.776         & 88.716*           & 80.455  & \textbf{89.889} \\
\multicolumn{1}{c|}{}                           & smooth(\%)↓ & 74.815  & \textbf{0.006} & 0.767            & 9.113   & 0.179*          \\ \hline
\end{tabular}
\end{table}

\subsubsection{Implementation Details.}
The implementation of our code uses PyTorch 3.9.0 \cite{paszke2019pytorch}, and the experiments were conducted on a single Nvidia A100 GPU. We use the Adam optimizer \cite{kingma2014adam} on a mini-batch of size $3$ with the initial learning rate $lr = 0.0002$ and the momentum parameters $\beta_1 = 0.5$ and $\beta_2 = 0.999$. We train our model for $200$ epochs and activate linear learning rate decay after every $20$ epochs. We implement the registration network by U-NET \cite{ronneberger2015u} with residual connections in the encoder and output paths. The number of INN blocks in the translation model is $8$. In the total loss function, the parameters $ \lambda_\alpha,  \lambda_\beta, \lambda_\gamma \text{ and } \lambda_\delta$ are asigned as $1, 1, 0.1,$ and $10$, respectively. The boundary $b$ in the barrier NMI loss is set to $5$.

\subsubsection{Comparison Methods.}
To demonstrate the superiority of the proposed INNReg method, we compare it with four widely-used multi-modal image registration methods, including the classical deep learning-based method Voxelmorph \cite{balakrishnan2019voxelmorph} and three translation-based methods: Arar et al. \cite{arar2020unsupervised}, RegGAN \cite{kong2021breaking}, and Chen et al. \cite{ijcai2022p117}. In Voxelmorph, we use the local cross-correlation as the similarity loss. For a fair comparison, all comparison methods are re-trained using the same training dataset as ours.

\begin{figure}[t]
  \centering
  \includegraphics[width=1\textwidth]{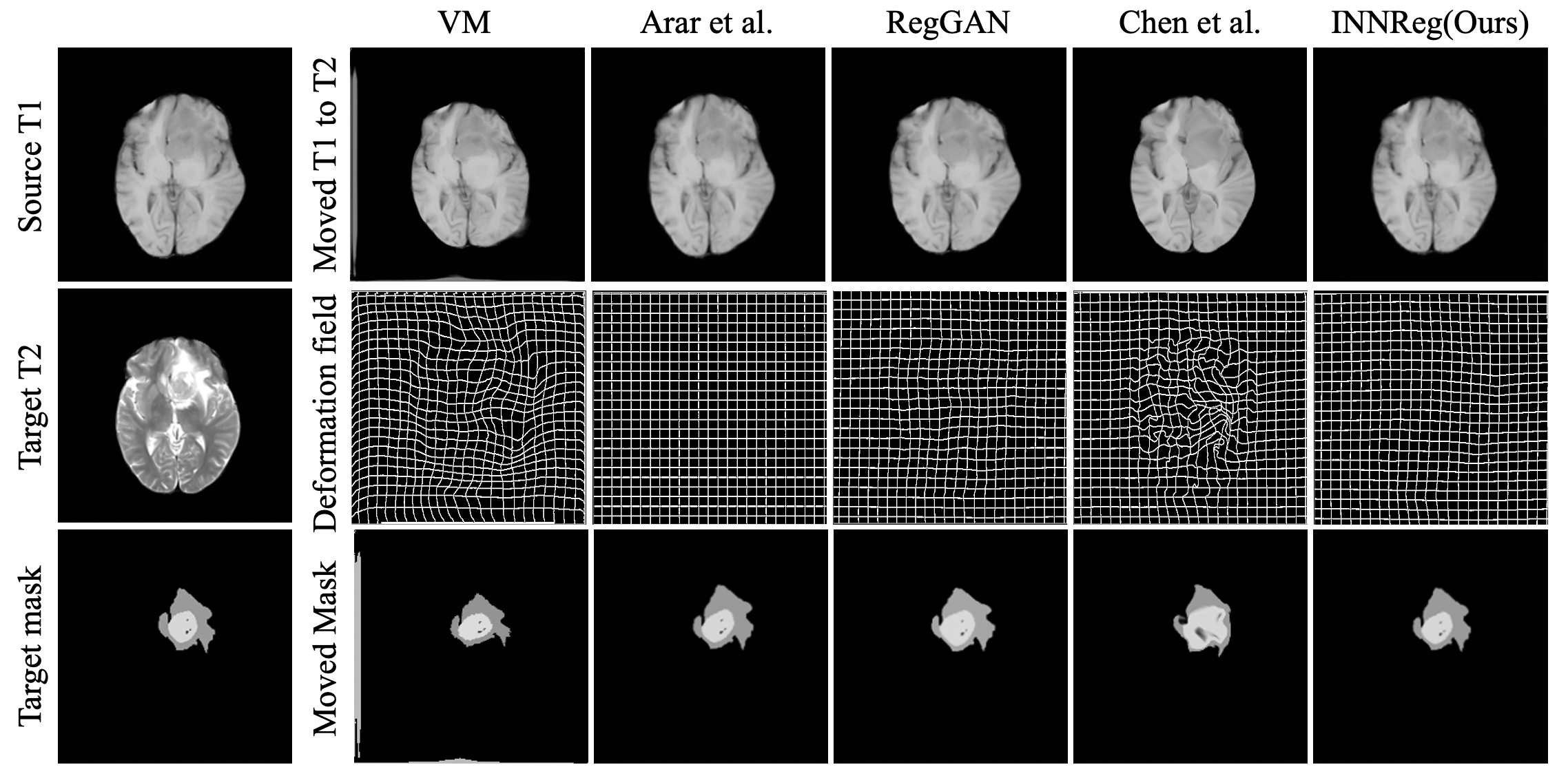}
  \caption{Visualization results of INNReg against other four methods on T1/T2 images. 
  The first row shows the warped source image $x(\phi)$, and the second displays the corresponding deformation field $\phi$. The last row exhibits the warped segmentation mask of the source image.
  }
  \label{fig_r1}
\end{figure}

\subsubsection{Evaluation Metrics.}
We evaluate the registration performance in terms of image similarity, mask similarity, and diffeomorphism. A higher image and mask similarity and a smoother diffeomorphism indicate higher registration accuracy. To measure image similarity, we adopt two standard metrics, including Structural Similarity (SSIM) and Normalized Correlation Coefficient (NCC), where higher SSIM and NCC indicate higher similarity in image level. Our quantitative metrics for measuring mask similarity include the Dice score \cite{dice1945measures} and the 95\% maximum Hausdorff distance (HD95) \cite{huttenlocher1993comparing}, which measure the degree of overlap between two regions. The higher Dice score and HD95 depict a higher mask similarity. We measure diffeomorphism by the smoothness of the deformation field $\phi$, which can be quantified by the determinant of its Jacobian matrix $J_{\phi}$. Specifically, we count the percentage of the pixels $p$ in each image such that $\left | J_{\phi}(\mathbf{p}) \right | \le 0$, and a smaller percentage implies higher smoothness of the deformation field, which further causes smoother diffeomorphism.

\subsection{Results on T1/T2 Dataset}
The quantitative registration results on the T1/T2 datasets are summarized in Table \ref{tab:compare}. Based on these results, we compare the proposed INNReg network with four other methods in terms of five evaluation metrics. From Table \ref{tab:compare}, we observe that among all the five evaluation metrics, INNReg achieves the best in terms of three of them (SSIM, NCC, Dice) and the second best in the remaining two (HD95, smooth). Note that although Arar et al. achieves the best smoothness, it is caused by the deformation that occurs in its translation network rather than the anticipated registration network, leading to a negligible deformation field. Consequently, INNReg outperforms the other comparison methods.

\begin{figure}[t]
  \centering
  \includegraphics[width=0.9\textwidth]{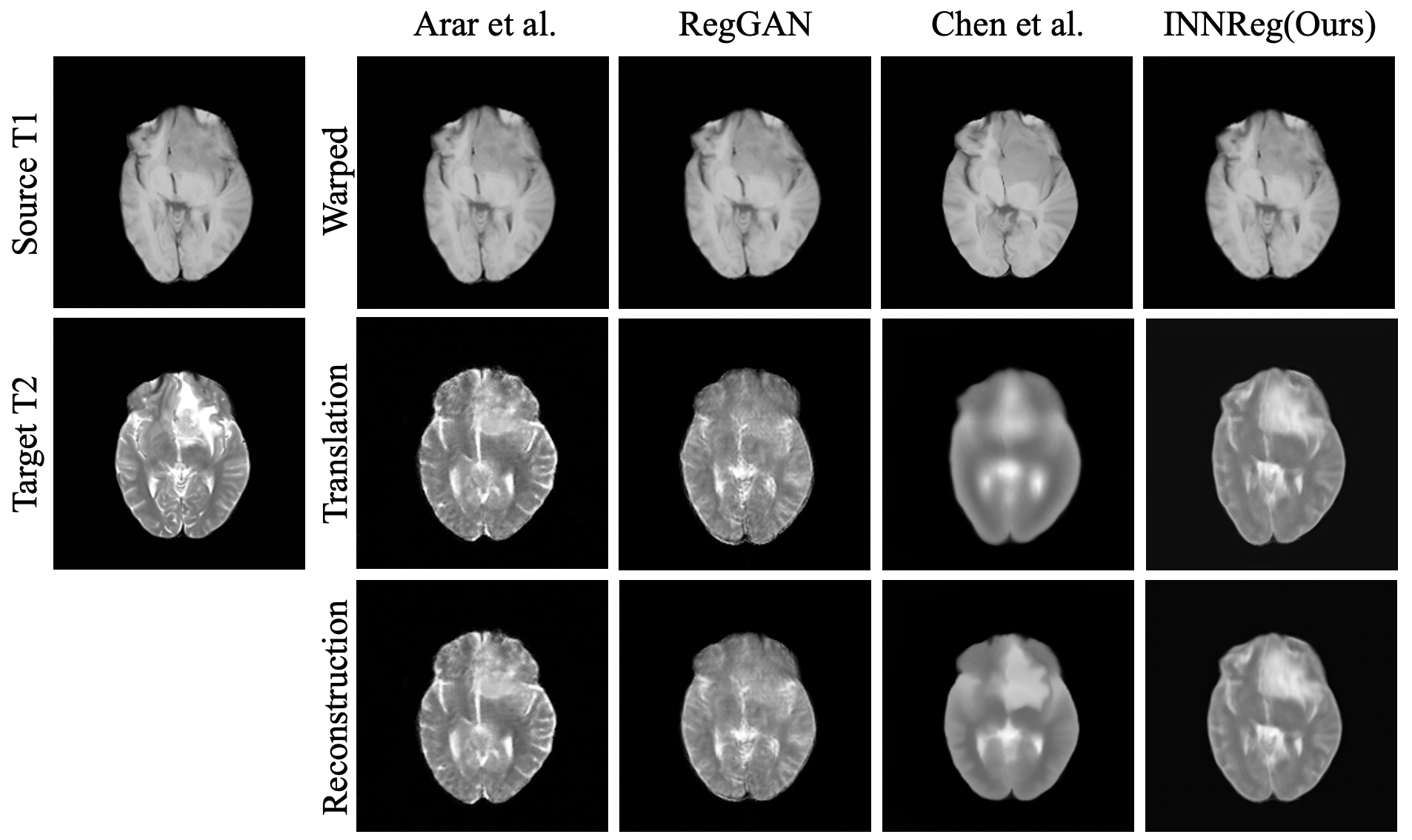}
  \caption{Visualization results of INNReg against the other three translation-based methods on T1/T2 images. The first row shows the warped source T1 image $x(\phi)$, and the second shows the translated results $\text{fake}\_y$. The last row displays the reconstructed results $\text{fake}\_y(\phi)$.
  }
  \label{fig_1_2}
\end{figure}

Fig. \ref{fig_r1} visualizes the multi-modal image registration results on the T1/T2 dataset. Specifically, it illustrates an instance of the warped source image, the corresponding deformation fields, and the warped source segmentation mask. By comparing the warped source images and the target image, we observe that INNReg achieves superior alignment between the source and target images. Furthermore, INNReg generates a smooth deformation field and the most accurate mask movement towards the target.

Fig. \ref{fig_1_2} displays not only the final warped image but also the intermediate results, which include the translated image ($\text{fake}\_y$ in Fig. \ref{fig_outline}) and the reconstruction result ($\text{fake}\_y(\phi)$ in Fig. \ref{fig_outline}). From Fig. \ref{fig_1_2}, we find a promising reason for the excellent registration performance of INNReg over the other translation-based methods -- both Arar et al and RegGAN exhibit significant deformation in the translated results, which notably impacts registration accuracy; The method proposed by Chen et al. does not have this issue but the quality of the translated image is unsatisfactory. Compared to these three methods, the translated image generated by our INNReg avoids both issues, and the superior translation results further lead to high-quality registration results. 

\subsection{Results on MRI/CT Dataset}
We also evaluate the effectiveness of the proposed INNReg on MRI/CT images. Due to the absence of segmentation masks, our evaluation focuses on the image similarity metrics SSIM and NCC, and the diffeomorphism of the deformation field. The quantitative results are displayed in Table \ref{tab:compare} and reveal that our method surpasses other compared methods in terms of SSIM and NCC. The smoothness of INNReg is inferior only to Arar et al., due to the negligible deformation in the translation process of Arar et al. as previously discussed.

\begin{figure}[t]
  \centering
  \includegraphics[width=0.9\textwidth]{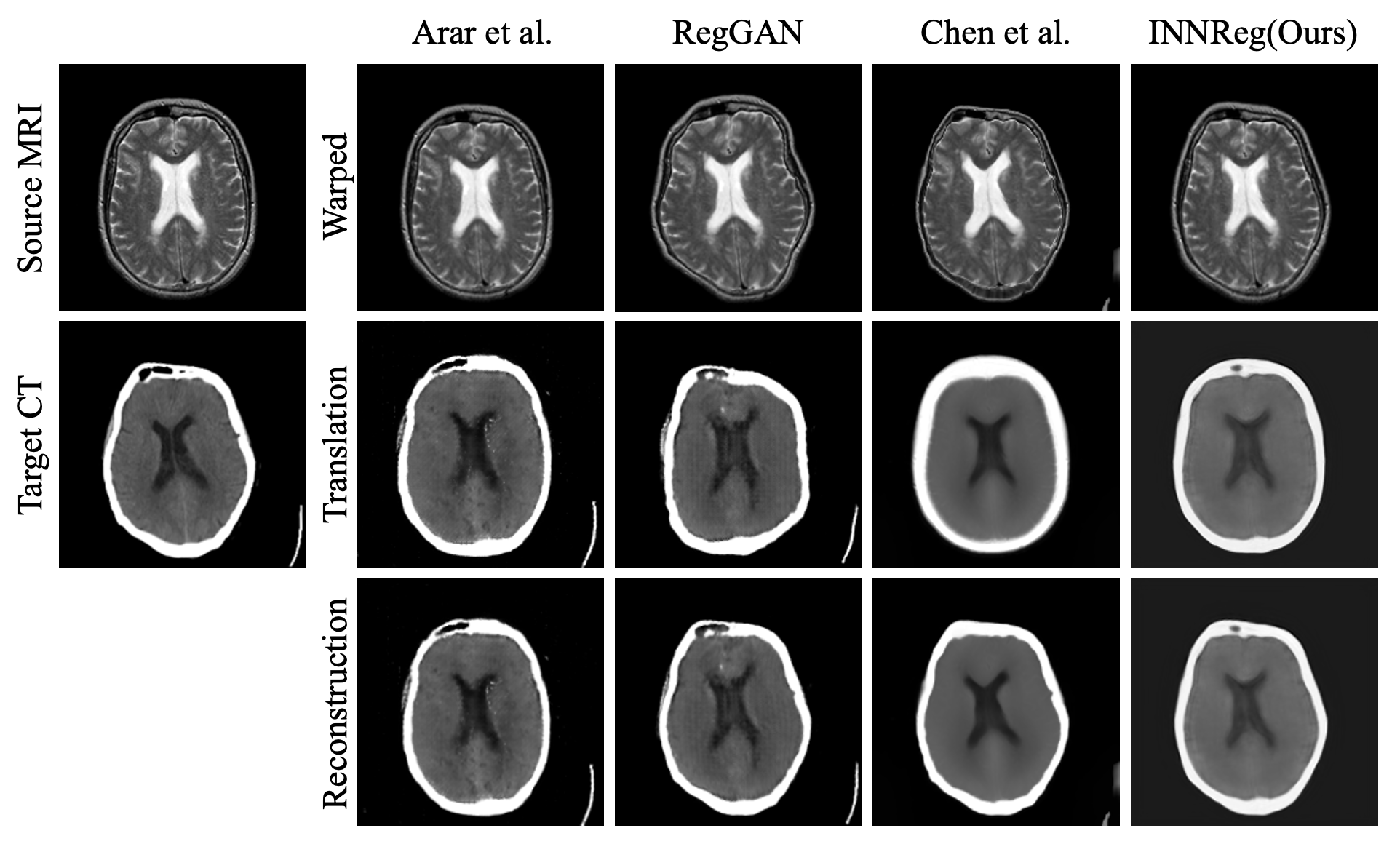}
  \caption{Visualization results of INNReg against the other three translation-based methods on MRI/CT images. The first row depicts the warped source MRI image $x(\phi)$, and the second shows the translated results $\text{fake}\_y$. The last row shows the reconstructed results $\text{fake}\_y(\phi)$.
  }
  \label{fig_R2}
\end{figure}

Fig. \ref{fig_R2} visualizes the warped source image, the translated image, and the reconstructed image in the experiments. It shows that INNReg achieves high-accuracy image registration on the MRI/CT image pairs. Additionally, we observe that our translation result remains highly geometry consistent with the source image, and the reconstructed image closely resembles the target image.

\subsection{Ablation Study}
\setlength{\tabcolsep}{3pt}
\begin{table}[tb]
  \caption{Ablation experiment results on T1/T2 images. \textbf{Bold} indicates the best value.}
  \label{tab:ablation}
  \centering
\begin{tabular}{c|ccccc}
\hline
Methods                & SSIM(\%)↑      & NCC(\%)↑        & Dice(\%)↑       & HD95↓              & smooth(\%)↓    \\ \hline
Base(A)            & 87.621         & 98.192          & 83.026          & 139.55722          & 0.128          \\
A+Cross translation(B) & 88.201         & 98.379          & 83.469          & 139.01316          & 0.115          \\
AB+NMI(C)              & 86.191         & 97.829          & 82.153          & 137.67901          & 0.135          \\
AB+Barrier NMI(D)      & \textbf{90.84} & 98.726          & 86.84           & 136.84976          & \textbf{0.095} \\
ABD+Attention(Ours)    & 90.555         & \textbf{98.746} & \textbf{87.117} & \textbf{101.37179} & 0.100          \\ \hline
\end{tabular}
\end{table}

In this subsection, we perform a series of ablation experiments to examine the impact of the INN cross-translation loss, the barrier NMI loss, and the DDWConv-based local attention on the registration accuracy of the proposed INNReg.  These experiments are conducted using T1/T2 images, and the results are summarised in Table \ref{tab:ablation}, in which the Base model refers to the version of INNReg stripped of the INN cross-translation loss, the barrier NMI loss, and DDWConv-based local attention. We evaluate the effectiveness of each component by comparing the performance of the models with and without it. From Table \ref{tab:ablation}, we make the following observations: First, the cross-translation and the barrier NMI loss enhance the registration results in all five metrics; Second, the barrier NMI loss is much more effective compared to NMI in terms of all the metrics; Third, although the DDWConv-based attention slightly deteriorates SSIM and smooth, it significantly improves the segmentation metrics, including Dice and HD95.

\section{Conclusion}

In this study, we introduced INNReg, an innovative unsupervised translation-based method for multi-modal image registration. Central to our approach is the exploitation of the Invertible Neural Network's (INN) reversible nature, enabling precise modality mapping and geometric consistency in I2I translation. By integrating a dynamic depthwise convolution-based local attention mechanism, our method effectively enhances feature extraction during the translation process. Furthermore, we proposed a novel barrier NMI loss function to rigorously constrain the registration process, which avoids the accuracy degradation of the registration network caused by severe deformation in the translation network. The efficiency and rationality of INNReg were validated across two public multi-modal medical image datasets, including T1/T2 MRI and MRI/CT image pairs. The experiment results demonstrated that INNReg well preserves geometry in translation and achieves high registration accuracy. In the future, we will explore further potentials of INNReg in multi-modal image translation tasks.

\section*{Acknowledgements}
We specifically acknowledge Dr Jinming Duan for his supervisory guidance and intellectual contributions to the first study, where he is not listed as a co-author of the formal publication because of an unintentional omission in the original submission.

%
%
\bibliographystyle{splncs04}
\bibliography{egbib}
\end{document}